\makeatletter\def\graphicscache@inhibit{true}\makeatother
\documentclass[runningheads]{llncs}
\usepackage{graphicx}

\usepackage{amsmath} %
\usepackage{amssymb}

\IfFileExists{graphicscache.sty}{\usepackage{graphicscache}}

\usepackage{tikz} %
\usetikzlibrary{arrows}
\usetikzlibrary{positioning,calc}
\usetikzlibrary{decorations.pathreplacing}
\usetikzlibrary{decorations.markings}
\usetikzlibrary{fit}
\usetikzlibrary{shapes.callouts}
\usetikzlibrary{shapes.geometric}
\usetikzlibrary{shapes.arrows}
\usetikzlibrary{matrix}
\usetikzlibrary{calendar}

\usepackage{pstricks}
\usepackage[capitalize]{cleveref}

\usepackage{multirow}

\usepackage{booktabs}
\usepackage{threeparttable}

\usepackage{fixme}
\fxsetup{status=draft}
\usepackage{float}

\tikzstyle{features}=[rectangle,draw,fill=green!80,rounded corners, minimum width=80pt, align=left]
\tikzstyle{model}=[rectangle,draw,fill=yellow!50, minimum width=70pt, minimum height=40pt, align=center]
\tikzstyle{group}=[rectangle, draw, dashed, minimum width=110pt, minimum height=130pt]

\usepackage{pgfplotstable}

\usepackage[numbers,sort&compress]{natbib}
\begin{document}

\title{ConvPoseCNN2: Prediction and Refinement of Dense 6D Object Poses}

\author{Arul Selvam Periyasamy, Catherine Capellen, Max Schwarz, and Sven Behnke}

\authorrunning{A. S. Periyasamy,  C. Capellen,  M. Schwarz, and  S. Behnke}

\institute{Autonomous Intelligent Systems, University of Bonn, Germany.\\
\email{periyasamy@ais.uni-bonn.de}}
\maketitle              %
\begin{abstract}
Object pose estimation is a key perceptual capability in robotics.
We propose a fully-convolutional extension of the PoseCNN method,
which densely predicts object translations and orientations.
This has several advantages such as improving the spatial resolution of the
orientation predictions---useful in highly-cluttered arrangements,
significant reduction in parameters by avoiding full connectivity,
and fast inference.
We propose and discuss several aggregation methods for dense orientation
predictions that can be applied as a post-processing step, such as
averaging and clustering techniques.
We demonstrate that our method achieves the same accuracy as PoseCNN
on the challenging YCB-Video dataset and provide a detailed ablation study
of several variants of our method.
Finally, we demonstrate that the model can be further improved by
inserting an iterative refinement module into the middle of the network,
which enforces consistency of the prediction.

\end{abstract}

\section{Introduction}

6D object pose estimation is an important building block for many applications, such as robotic manipulation. While many objects can be grasped without precise pose information, there are many
tasks which require 6D pose estimates, for example
functional grasping of tools and assembly.
Such tasks routinely come up in industrial applications, as evidenced by the Amazon Picking \& Robotics Challenges 2015-2017, where pose estimation played a key role for difficult objects,
but can also appear in semi-unstructured environments, as in home and assistance robotics.

State-of-the-art pose estimation methods predominantly use CNNs for 6D object pose estimation from RGB(-D) images. One of the notable features of these methods is the joint learning of multiple simultaneous tasks such as object detection, semantic segmentation, and object pose estimation. Although 6D object pose estimation from \mbox{RGB-D} images is an active area of research, for the sake of brevity, we focus on monocular, i.e. RGB only methods. These methods can be broadly classified into two categories: direct regression methods, and 2D-3D correspondence methods. The direct regression methods estimate 6D pose directly from input images, for example in the form of a 3D vector (translation) and a quaternion (orientation). Examples of these methods include \citet{deep6dpose} and \citet{xiang2017posecnn}. In contrast, the correspondence-based methods
predict the projection of 3D points in the 2D image and recover the pose of the object by solving the Perspective-n-Point problem. These methods can be further classified into dense correspondence methods and keypoint-based methods. The dense correspondence methods~\citep{brachmann2016uncertainty,krull2015learning} predict the projected 3D coordinates of the objects per pixel while the keypoint-based methods~\citep{heatmaps,featuremapping,SSS6D,BB8} predict projection of 3D keypoints in the 2D image.

Since the CNN architecture we propose is closely related to PoseCNN~\citep{xiang2017posecnn}, a direct regression method, we review PoseCNN architecture in detail. PoseCNN learns to predict 6D pose objects jointly with semantic segmentation. The CNN uses a pretrained VGG~\citep{simonyan2014very} backbone followed by three branches to predict segmentation class probabilities, direction and distance to center, and orientation (represented as quaternions). The orientation prediction branch uses fully connected layers while the other two branches use fully convolutional layers. The orientation prediction branch takes a fixed size image crop as input. From the segmentation class probabilities, a crop containing a single object is extracted and resized to the fixed orientation prediction branch input size using a RoI pooling layer.

Introduced by \citet{girshick2015fast}, RoI pooling is a powerful mechanism
for scale normalization and attention and resulted in significant advancements
in object detection and related tasks. However, RoI pooling has drawbacks:
Especially in cluttered scenes, its cutting-out operation may disrupt flow of
contextual information. Furthermore, RoI pooling requires random access to
memory for the cutting-out operation and subsequent interpolation, which may
be expensive to implement in hardware circuits and has no equivalent in
the visual cortex~\citep{kandel2000principles}.

\Citet{redmon2016you} demonstrated that simpler, fully-convolutional architectures
can attain the same accuracy, while being tremendously faster and having fewer
parameters. In essence, fully-convolutional architectures can be thought of
as sliding-window classifiers, which are equivalent to RoI pooling with
a fixed window size. While the scale-invariance is lost, fully-convolutional
architectures typically outperform RoI-based ones in terms of model size
and training/inference speed.
When addressing the inherent example imbalances during training with a customized
loss function~\citep{lin2017focal}, fully-convolutional architectures reach
state-of-the-art performance in object detection.

Following this idea, we propose a fully convolutional architecture for pose estimation, which can \textit{densely}
predict all required information such as object class and transformation. If required, the
dense prediction can be post-processed and aggregated per object to obtain a final prediction.

Given the complex nature of the task, instead of directly predicting pose from the given RGB image of a scene, many approaches formulate pose estimation as an iterative refinement process:
Given an initial pose estimate and high quality 3D model of the objects, the objects are rendered as per current pose estimate, a refined pose that minimizes difference between the rendered and the observed image is predicted at each step and this step is repeated multiple times. \citet{li2018deepim} trained a CNN that takes RGB image and rendered image of a object as per the current pose estimate as input and predicts the a pose update that refines the current pose update in each step. This step is repeated until the pose update is negligible. \citet{periyasamy2019refining} used a differentiable renderer to compute pose updates to minimize difference between the rendered and the observed image. Unlike \citep{li2018deepim} that refines pose of single object at a time, \citep{periyasamy2019refining} refined poses for all objects in the scene at each iteration. \citet{krull2015learning} trained a CNN to predict a matching score---how similar are two images--- between the rendered and the observed image. The matching score was used to pick one best pose hypothesis among many available pose hypotheses. One prevalent characteristic among the pose refinement approaches is that refinement is done post prediction--refinement model and pose prediction model are decoupled. 
In contrast, our proposed iterative refinement module is built into the pose estimator. We enhance the ConvPoseCNN architecture from our previous work \citep{capellen2020convposecnn} with an iterative refinement module to learn representations suitable for both translation and orientation predictions instead of refining the predictions from the estimator.

In summary, our contributions include:
\begin{itemize}
 \item A network architecture and training regime for dense orientation prediction,
 \item aggregation \& clustering techniques for dense orientation predictions, and
 \item an iterative refinement module which increases prediction accuracy.
\end{itemize}

\begin{figure}[t]
\centering
\includegraphics[width=0.9\textwidth]{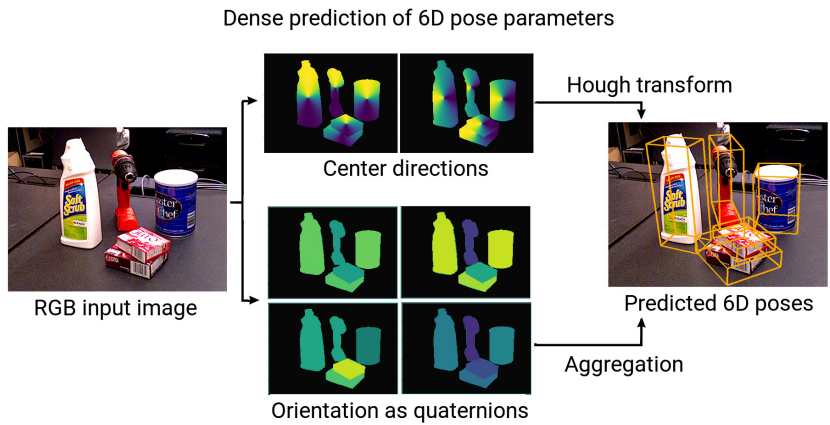}
\caption{Dense Prediction of 6D pose parameters inside ConvPoseCNN. The dense
 predictions are aggregated on the object level to form 6D pose outputs.
 Source:~\citet{capellen2020convposecnn}.}
\end{figure}

\begin{figure*}
\centering
\includegraphics[width=0.9\textwidth]{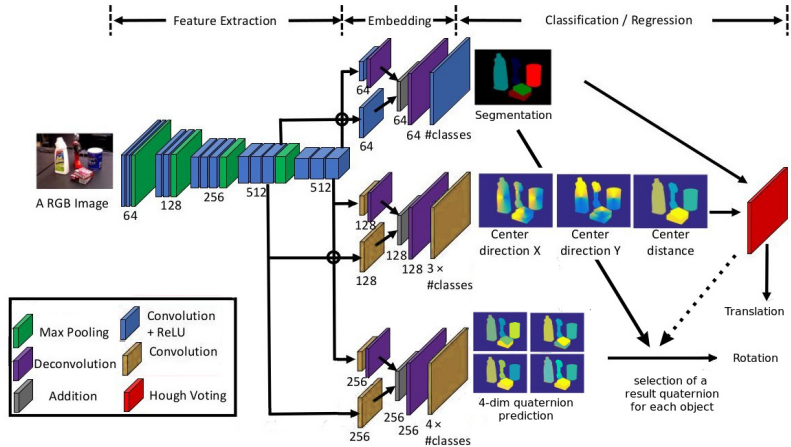}
\caption{Our ConvPoseCNN architecture for convolutional pose estimation.
During aggregation, candidate quaternions are selected according to
the semantic segmentation results or according to Hough inlier information.
Source:~\citet{capellen2020convposecnn}.}.
\label{pic:convolutionalnetwork}
\end{figure*}

\section{ConvPoseCNN}

We propose an extension of the PoseCNN~\citep{xiang2017posecnn} architecture.
The PoseCNN network is based on a VGG backbone
with three heads: One performing semantic segmentation, one densely
predicting object center directions in 2D and object depth, and finally a
RoI-Pooling branch with a fully connected head predicting one orientation
quaternion for each object.
Our proposed network keeps most of this structure, but replaces the orientation
prediction branch with a fully convolutional one, which estimates orientation
\textit{densely} (see \cref{pic:convolutionalnetwork}). The architecture of the new branch is modeled after the translation estimation
branch.

The dense translation prediction is post-processed during inference as by
\citet{xiang2017posecnn}: The 2D center predictions are fed into a Hough voting
layer, which aggregates them into center hypotheses. The predicted object
depth is averaged over all inliers. Finally, the 3D position can be computed
through ray projection using the camera intrinsics.

\subsection{Aggregation of Dense Orientation Predictions}
\label{method:aggregation}

Estimating the final orientation prediction from pixel-wise quaternion predictions
is not as straight-forward, however.
We investigate two different approaches for this purpose: averaging and clustering.

Quaternions corresponding to a rotation, by definition, have unit norm. But we do not enforce the quaternion predictions to be of unit norm explicitly during the ConvPoseCNN training.
Thus, before aggregating the dense predictions, we need to scale them to unit norm. Interestingly, we observe that the norm of the quaternion at a pixel prior to scaling corresponds to quality of the prediction. i.e. pixels in the feature-rich regions of the image have higher quaternion norm.
Exploiting this observation, we use the norm of the quaternion prediction $w = ||q||$ as an optional weighting factor in our aggregation step.
We extract the quaternions $q_1,...,q_n$ corresponding to an object using the segmentation predictions and average them following the the optimization scheme proposed by \citep{markley2007quaternion} using the norm $w_1,...,w_n$. The average quaternion $\bar{q}$ is given by

\begin{equation}
 \bar{q} = \text{arg} \min_{q \in \mathbb{S}^3} \sum_{i=1}^n w_i ||R(q) - R(q_i)||^2_F,
\end{equation}
where $R(q) - R(q_i)$ are the rotation matrices corresponding to the quaternions, $\mathbb{S}^3$ is the unit 3-sphere, and $||\cdot||_F$ is the Frobenius norm.
Note that quaternion to rotation matrix conversion eliminates any problems arising from the antipodal symmetry of
the quaternion representation. The exact solution to the optimization problem
can be found by solving an eigenvalue problem~\citep{markley2007quaternion}.
In case of multiple, overlapping instances of the same object class---here, the
predicted segmentation would not be enough to differentiate the
instances---we can additionally make use of the Hough voting procedure required
for translation estimation to separate the predictions into inlier sets for
each object hypothesis.

Averaging based aggregation schemes inherently may from suffer skewed results due to bad outlier predictions. Clustering based aggregation schemes should be less susceptible to outlier predictions.

We follow a weighted RANSAC clustering scheme as an alternative to averaging: For quaternions $Q=\{q_1, ..., q_n\}$ and their weights $w_1, ..., w_n$  associated with one object we repeatedly choose a random quaternion $\bar{q} \in Q$ with a probability proportional to its weight and then
determines the inlier set $\bar{Q}=\{q \in Q|d(q,\bar{q}) < t\}$, where $d(\cdot,\cdot)$ is the angular distance.
Finally, the $\bar{q}$ with largest $\sum_{q_i \in \bar{Q}} w_i$ is selected as the result quaternion.

\subsection{Iterative refinement}
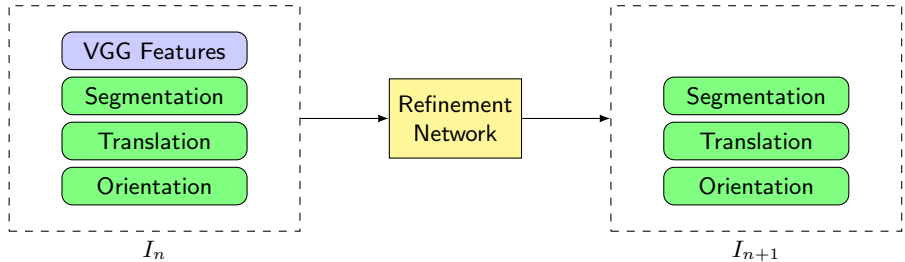
\begin{figure}
﻿\begin{tikzpicture}[
  font=\sffamily\footnotesize
]

\tikzstyle{model}=[rectangle,draw,fill=yellow!50, minimum width=50pt, minimum height=30pt, align=center]                                                                                                                                                                                                                      
\tikzstyle{features}=[rectangle,draw,fill=green!50,rounded corners, minimum width=70pt, align=center, minimum height=0.5cm,text depth=0cm] 
\tikzstyle{layers}=[rectangle,draw,fill=red!30,rounded corners, minimum width=50pt, align=center] 
\tikzstyle{group}=[rectangle, draw, dashed, minimum width=110pt, minimum height=130pt]

 \node at (0,0.9) [features,fill=blue!20] {VGG Features};
 \node at (0,0.3) [features] {Segmentation};
 \node at (0,-0.3) [features] {Translation};
 \node at (0,-0.9) [features] {Orientation};

 \node(cnn) at (4, 0)[model, ]{Refinement \\  Network};
 \node(input) at (0,0) [ group, minimum height=3cm] {};
 \node(output) at (8,0) [ group, minimum height=3cm] {};

\node[below=0cm of input] {$I_n$};
\node[below=0cm of output] {$I_{n+1}$};

 \node at (8,0.3) [features, ] {Segmentation};
 \node at (8,-0.3) [features,] {Translation};
 \node at (8,-0.9) [features,] {Orientation};
\node[name] (inputanchor)  at (1.8,2)[] {};
\node[name] (outputanchor)  at (6.2,2)[] {};

\draw[-latex] (cnn) -- (output);
\draw[-latex](input) -- (cnn);

\end{tikzpicture}
   \caption{Naive post-prediction iterative refinement of segmentation probabilities, translation predictions, and rotation predictions. The dense predictions (green) are refined using a small network, which can be applied repeatedly for further refinement. High-level features (blue) can be fed into the network to provide additional
  context information.}
  \label{fig:tikz:ir_final}
\end{figure}

\label[]{sec:ir}
During the prediction of 6D object poses, translation estimates and orientation estimates influence each other. Predicting translation and orientation components using separate branches as in ConvPoseCNN and PoseCNN does not allow the model to exploit the interdependence between translation and orientation estimates. This motivates in designing network architectures that can refine translation and orientation prediction iteratively to enable the network to model the interdependencies between the predictions. One naive way of doing pose refinement would be to perform refinement after prediction. To this end, we experimented with a simple three layered---three blocks of convolutional layer followed by ReLU activation---fully convolutional model to refine the predictions from ConvPoseCNN model iteratively.
At each step, segmentation, translation, and orientation predictions along with the features from the VGG backbone model are provided as input and a refined estimate is computed as depicted in Figure. \ref{fig:tikz:ir_final}. The final predictions are obtained after a small fixed number of iterations. We call this approach post-prediction iterative refinement.

\begin{figure}
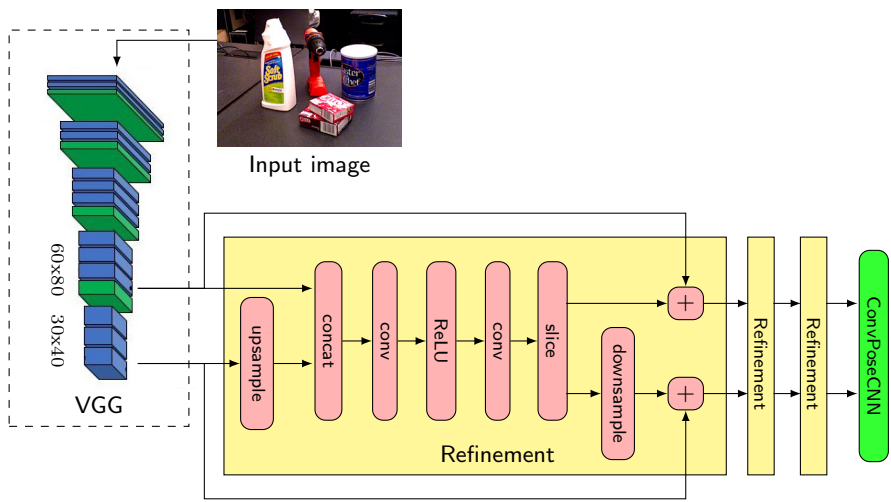

﻿\begin{tikzpicture}[
  font=\sffamily\small,
  con/.style={-latex, black}
]

\tikzstyle{model1}=[rectangle,draw,fill=yellow!50, minimum width=190pt, minimum height=90pt, align=center]                                                                                                                                                                                                                      
\tikzstyle{feat}=[rectangle,draw,fill=blue!30,rounded corners, minimum width=50pt, align=left] 

\tikzstyle{layers}=[rectangle,draw,fill=red!30,rounded corners, minimum width=50pt, align=left] 

\node[inner sep=0pt] (inimg) at (3,4)
    {\includegraphics[width=.2\textwidth]{figures/ycb_input.png}};
\node[below=0cm of inimg] {Input image};

\node[inner sep=0pt, , rotate=-90, anchor=north] (vggmodel) at (1.5, 2)
    {\includegraphics[width=.35\textwidth]{figures/VGG.png}};

 \node(cnn) at (5.2, 0.3)[model1, ]{};
 \node(cnn2) at (9, 0.3)[model1, minimum width=10pt ]{};
 \node(cnn3) at (9.7, 0.3)[model1, minimum width=10pt ]{};
 \node(up1) at (9, 0.2) [rotate=-90]  {\scriptsize Refinement};
 \node(up1) at (9.7, 0.2) [rotate=-90]  {\scriptsize Refinement};

 \node(input) at (0.2,2.) [ group, minimum height=150pt, minimum width=68pt] {};

\node[name] (vgg4)  at (0. 6, 1.2)[] {};
\node[name] (vgg41)  at (1.3, 1.1)[] {};
\node[name] (vgg42)  at (1.3, 2.4)[] {};
\node[name] (vgg421)  at (1. 2, 2.25)[] {};
\node[name] (vgg43)  at (8.15, 2.25)[] {};
\node[name] (vgg431)  at (8. , 2.4)[] {};

\node[name] (vgg5)  at (0.6, 0.2)[] {};
\node[name] (vgg51)  at (1. 3, 0.35)[] {};
\node[name] (vgg52)  at (1.3, -1.78)[] {};
\node[name] (vgg521)  at (1.2, -1.65)[] {};
\node[name] (vgg53)  at (8. 15, -1.65)[] {};
\node[name] (vgg531)  at (8. , -1.65)[] {};

\node[name] (cat1)  at (3.2, 1.2)[circle] {};
\node[name] (cat2)  at (3.2, 0.2)[circle] {};
\node[name] (slice1)  at (6.25, 1.)[circle] {};
\node[name] (slice2)  at (6.25, -0.2)[circle] {};

\node[name] (in1)  at (8.95, 1.)[] {};
\node[name] (in2)  at (8.95, -0.2)[] {};

 \node(up1) at (2.3, 0.2) [layers, rotate=-90]  {\scriptsize upsample};
 \node(cat) at (3.25, .5) [layers, rotate=-90, minimum width=60pt] {\scriptsize concat};
 \node(conv) at (4, .5) [layers, rotate=-90, minimum width=60pt] {\scriptsize conv};
 \node(rel) at (4.75, .5) [layers, rotate=-90, minimum width=60pt] {\scriptsize ReLU};
 \node(conv2) at (5.5, .5) [layers, rotate=-90, minimum width=60pt] {\scriptsize conv};

 \node(slice) at (6.23, .5) [layers, rotate=-90, minimum width=60pt] {\scriptsize slice};
 \node(down1) at (7.1, -0.2) [layers, rotate=-90]  {\scriptsize downsample};
  \node(resadd) at (8, -0.2) [layers, minimum width=1pt] {$+$};
  \node(resadd1) at (8, 1) [layers, minimum width=1pt] {$+$};

\draw[con] (vgg5) -- (up1);
\draw[con] (vgg5) -- ++(1,0) -| ++(0,-1.8) -| (resadd);

\draw[con] (vgg4) -- (cat1);
\draw[con] (vgg4) -- ++(1,0) -| ++(0,1.0) -| (resadd1);

\draw[con] (up1) -- (cat2);
\draw[con] (cat) -- (conv);
\draw[con] (conv) -- (rel);
\draw[con] (rel) -- (conv2);
\draw[con] (conv2) -- (slice);

\draw[con] (slice1) -- (resadd1);
\draw[con] (slice2) -- (down1);

\draw[con] (down1) -- (resadd);

\draw[con] (resadd) -- (in2);
\draw[con] (resadd1) -- (in1);

\node[name] (test1)  at (4, 2)[] {};
\node[name] (test2)  at (4, 4)[] {};

\draw[con] ($(inimg.west)+(0,0.5)$) -| (vggmodel.west);

 \node at (10.5, .3) [features, rotate=-90] {\scriptsize ConvPoseCNN};
\node[name] (in11)  at (9.05, 1.)[] {};
\node[name] (in12)  at (9.05, -0.2)[] {};
\node[name] (in21)  at (9.65, 1.)[] {};
\node[name] (in22)  at (9.65, -0.2)[] {};

\draw[con] (in11) -- (in21);
\draw[con] (in12) -- (in22);

\node[name] (in31)  at (9.75, 1.)[] {};
\node[name] (in32)  at (9.75, -0.2)[] {};
\node[name] (in41)  at (10.4, 1.)[] {};
\node[name] (in42)  at (10.4, -0.2)[] {};

\draw[con] (in31) -- (in41);
\draw[con] (in32) -- (in42);

\node at (0.2, -0.35)[] {\small VGG};
\node at (5.5, -1)[] {\small Refinement};

\node at (-0.35, 1.45)[font=\large,rotate=-90] {\scriptsize 60x80};
\node at (-0.35, 0.5)[font=\large,rotate=-90] {\scriptsize 30x40};

\end{tikzpicture}
   \caption{%
    Pre-prediction iterative refinement of the features extracted from the VGG network.
    The refined features are then fed into the ConvPoseCNN head networks.
    Note that there is only one set of weights for the refinement module,
    i.e. it is applied iteratively.
  }
  \label{fig:tikz:middle}
\end{figure}

However, naive post-prediction refinement might be a challenging task because the predictions might be in a suitable form for a simple three layer model. To address this concern, we experimented with a pre-prediction iterative refinement of intermediate representation  shown in Figure \ref{fig:tikz:middle}. The features from the pretrained VGG backbone model ae refined before providing them as input to ConvPoseCNN network enabling ConvPoseCNN model to learn joint intermediate representations suitable for both translation and orientation predictions. The refinement module is akin to residual blocks in ResNet architecture (\citet{ren2015faster}). Each iteration refinement module computes $\Delta(x)$ that is added to the input with the use of skip connections.
$$ f^{i+1}(x)= f^{i}(x) + \Delta(x) $$

In detail, refinement blocks takes two set of features maps ${F_{A}}$, and ${F_{B}}$ each of dimension $512$x$60$x$80$, and $512$x$30$x$40$ respectively as input. ${F_{B}}$  is upsampled with transposed convolution and concatenated with ${F_{A}}$. 
The resulting  $1024$x$60$x$80$ is passed through a sequence of convolutional, ReLU, and convolutional layers. All the convolutions have a window size of 3 and stride of 1. Zero padding of one pixels is applied to maintain the spatial resolution of the features.
 
Then the features are split to two equal parts. One of them is downsampled. Thus we arrive at $\Delta{F_{A}}$ and $\Delta{F_{B}}$ having same spatial dimensions as of  ${F_{A}}$, and ${F_{B}}$ respectively. Using skip connections $\Delta{F_{A}}$ and $\Delta{F_{B}}$ and added to ${F_{A}}$, and ${F_{B}}$ respectively.

\section{Evaluation}

\subsection{Dataset}
We evaluate our method on the YCB-Video dataset~\citep{xiang2017posecnn}. The dataset contains 133,936 images of VGA resolution (640$\times$480) extracted from 92 video sequences. Each image contains a varying number of objects selected from a larger set of 21 objects, some of which exhibit symmetry or are texture-poor. The first frame of each of the 92 video sequences was manually annotated with ground truth pose information, while the rest of the frames were automatically annotated by tracking the camera pose via SLAM. In each sequence, the objects are arranged in various spatial configurations resulting in varying degrees of occlusions making it a challenging dataset. High quality 3D models and downsampled point models containing 2620 points each are made available. The real images are supplemented with  simple synthetic renderings of the object models.

\subsection{Training procedure}
\label{sub:training}

We implemented ConvPoseCNN using PyTorch \citep{paszke2017automatic} framework. Except for the novel dense orientation estimation branch, we based our implementation on the openly available PoseCNN implementation \footnote{\label{note:posecnn}\url{https://github.com/yuxng/PoseCNN}}. The openly available official implementation has minor changes compared to the model described by  \citep{xiang2017posecnn}. We noted that these minor design choices were helpful and incorporated them in our implementation as well. Additionally, we implemented Hough voting layer---non differentiable layer that computes inlier pixels---using Numba \citep{numba}. Although Numba is CPU only, the Hough Voting layer implementation with Numba is faster than other GPU implementations. 

The segmentation and translation branches of ConvPoseCNN are trained with the standard pixelwise negative log-likelihood (NLL) and L2 loss respectively. The depth component of the translation branch is of a smaller scale compared to the other two components. To balance this discrepancy we scale the depth component loss by 100.

We use the ShapeMatch loss ($\text{SMLoss}$) proposed by \citep{xiang2017posecnn} to train the orientation branch of ConvPoseCNN. $\text{SMLoss}$ handles objects with and without symmetry using two different loss definitions as follows.
\begin{equation}
 \text{SMLoss}(\tilde{q},q) = \begin{cases}
                                       \text{SLoss}(\tilde{q},q), & \text{if object is symmetric,} \\
                                       \text{PLoss}(\tilde{q},q), & \text{otherwise.}
                                      \end{cases},
\end{equation}

Given a set of 3D points $\mathbb{M}$, where
 m = $|\mathbb{M}|$ and $R(q)$ and $R(\tilde{q})$ are the rotation matrices corresponding to ground truth and estimated quaternion, respectively, and PLoss and SLoss are defined as follows:
\begin{alignat}{2}
\text{PLoss}(\tilde{q},q) &= \frac{1}{2m} \sum_{x\in \mathbb{M}} || R(\tilde{q})x - R(q)x||^2, \\
\text{SLoss}(\tilde{q},q) &= \frac{1}{2m} \sum_{x_1\in \mathbb{M}}\min_{x_2\in\mathbb{M}} || R(\tilde{q})x_1 - R(q)x_2||^2.
\end{alignat}
Similar to the ICP objective, SLoss does not penalize rotations of symmetric objects that lead to equivalent shapes.

During the training phase, computing $\text{SMLoss}$ per pixels is computationally infeasible. Thus, we resort to aggregate dense predictions for each object before computing loss functions. We experimented with the aggregation mechanisms discussed in \cref{method:aggregation} and observed poor convergence. We hypothesize that this could be because of weighting quaternions with their norm before aggregation results in pixels with smaller quaternion prediction norm receiving smaller gradients. 
Empirically, we found that using simple numerical averaging to arrive at $\tilde{q}$ alleviates the issue of uneven gradient distribution and contributes to convergence of the training process. Additionally, numerical averaging is computationally less expensive.

Alternatively, we also experimented with training the orientation branch with pixel-wise $\text{L2}$ loss and $\text{QLoss}$ \citep{billings2018silhonet}

For two quaternions $\bar{q}$ and $q$ it is defined as:
\begin{equation}
\text{QLoss}(\bar{q}, q) = \log(\epsilon + 1 - |\bar{q}\cdot q|),
\end{equation}
where $\epsilon$ is introduced for stability. QLoss is designed to handle the quaternion symmetry.

The final loss function used during training is, similarly to PoseCNN, a linear
combination of segmentation, translation, and orientation loss:
\begin{equation}
L = \alpha_{\text{seg}}L_{\text{seg}} + \alpha_{\text{trans}}L_{\text{trans}} + \alpha_{\text{rot}}L_{\text{rot}}.
\end{equation}

where
$\alpha_{\text{seg}}$, $\alpha_{\text{trans}}$, are set to 1. $\alpha_{\text{rot}}$ is set to 1 and 100 in the case of $\text{L2}$ loss and $\text{QLoss}$, and $\text{SMLoss}$ respectively.
We train ConvPoseCNN model using SGD with learning rate 0.001 and momentum 0.9.

\subsection{Evaluation Metrics}

We report area under the accuracy curve (AUC) metrics AUC-P and AUC-S for varying area threshold between 0 and 0.1m on ADD and ADD-S metrics as introduced along with YCB-Video Dataset \citep{xiang2017posecnn}. ADD is average distance between corresponding points of the 3D object model in predicted and ground truth pose. ADD-S is the average of distance between each 3D point in predicted pose to the closest point in ground truth pose.  ADD-S penalizes objects with symmetry less than ADD metric.

\subsection{Results}

\subsubsection{Prediction Averaging}

To aggregate the dense pixel-wise predictions into a single orientation estimate, we use weighted quaternion averaging \citep{markley2007quaternion}. In the case of ConvPoseCNN, there are two possible sources of the pixel-wise weighting: segmentation score, and predicted quaternion norm. In \cref{tab:resultspixell2}, we show the comparison between the two weighting schemes. The norm weighting showed better results than both no averaging and using segmentation score as weighting. This suggests the predictions with smaller norms are less precise. Encouraged by this observation, we experimented further with pruning the predictions before aggregation  We sorted the predictions based on the norm and pruned varying percentile number ($\lambda$) of them.  

\cref{tab:resultsl2pruned} shows results of pruning with percentile ranging from 0 (no pruning) to 1 (extreme case of discarding all but one prediction). Pruning improves the results by a small factor overall but considerably for the objects with symmetry. This can be explained by the fact that averaging shape-equivalent orientations might result in an non-equivalent orientation and thus averaging schemes are not suitable for handling objects with symmetry.

\begin{table}[H]
  \centering
  \begin{threeparttable}
  \caption{Weighting strategies for ConvPoseCNN L2}
  \label{tab:resultspixell2}\setlength{\tabcolsep}{12pt}
  \begin{tabular}{lcccc}
  \toprule
           Method           & \multicolumn{2}{c}{ 6D pose\tnote{1} }    & \multicolumn{2}{c}{Rotation only}  \\
  \cmidrule(lr){2-3}\cmidrule(lr){4-5}
                        & AUC P          & AUC S          & AUC P           & AUC S      \\ \midrule
                  PoseCNN\tnote{2}                      & 53.71          & 76.12          & \textbf{78.87}         & \textbf{93.16} \\ \midrule
  unit weights                      & 56.59 & 78.86 & 72.87 & 90.68 \\
  norm weights             & \textbf{57.13} & \textbf{79.01} & 73.84 & 91.02 \\
  segm. weights & 56.63 & 78.87 & 72.95 & 90.71 \\
       \bottomrule
  \end{tabular}
  \begin{tablenotes}\footnotesize
   \item [1] Following \citet{xiang2017posecnn}.
   \item [2] Calculated from the published PoseCNN model.
  \end{tablenotes}
  Source:~\citet{capellen2020convposecnn}.
  \end{threeparttable}
  \end{table}

  \begin{table}[H]
  \centering
  \begin{threeparttable}
  \caption{Quaternion pruning for ConvPoseCNN L2}
  \label{tab:resultsl2pruned}\setlength{\tabcolsep}{12pt}
  \begin{tabular}{lcccc}
  \toprule
           Method           & \multicolumn{2}{c}{ 6D pose\tnote{1}}    & \multicolumn{2}{c}{Rotation only}  \\
  \cmidrule(lr){2-3}\cmidrule(lr){4-5}
                        & AUC P          & AUC S          & AUC P           & AUC S      \\
  \midrule
                  PoseCNN                      & 53.71          & 76.12          & \textbf{78.87}         & \textbf{93.16} \\ \midrule
  pruned(0)      & 57.13          & 79.01          & 73.84 & 91.02 \\
  pruned(0.5)    & \textbf{57.43} & 79.14          & 74.43 & 91.33 \\
  pruned(0.75)   & \textbf{57.43} & 79.19          & 74.48 & 91.45 \\
  pruned(0.9)    & 57.37          & \textbf{79.23} & 74.41 & 91.50 \\
  pruned(0.95)   & 57.39          & 79.21          & 74.45 & 91.50 \\
  single         & 57.11          & 79.22          & 74.00 & 91.46 \\ \bottomrule
  \end{tabular}
  \begin{tablenotes}
   \item[1] Following \citet{xiang2017posecnn}.
  \end{tablenotes}
  Source:~\citet{capellen2020convposecnn}.
  \end{threeparttable}
  \end{table}

\subsubsection{Prediction Clustering}

As an alternative to averaging schemes we experimented with RANSAC-based clustering schemes where we chose a quaternion at random and cluster the other quaternions into inliers and outliers based on the angular distance between corresponding rotations as the threshold. We repeat the process 50 times and select the quaternion prediction with the maximum inlier count. As opposed to the L2 distance in quaternion space, angular distance function invariant to the antipodal symmetry of the quaternion orientation representation. The results are shown in  \cref{tab:resultsransacl2}. Similar to averaging schemes, weighted variant of RANSAC performs better than non-weighted variants. Overall, clustering schemes outperform averaging schemes slightly on AUC S metric but perform slightly worse on AUC P. This is expected as the clustering schemes can handle object symmetries well.

\begin{table}[H]
  \centering
  \begin{threeparttable}
  \caption{Clustering strategies for ConvPoseCNN L2}
  \label{tab:resultsransacl2}\setlength{\tabcolsep}{12pt}
  \begin{tabular}{lcccc}
  \toprule
           Method           & \multicolumn{2}{c}{6D pose}    & \multicolumn{2}{c}{Rotation only}  \\
  \cmidrule(lr){2-3}\cmidrule(lr){4-5}
                        & AUC P          & AUC S          & AUC P           & AUC S      \\ \midrule
                  PoseCNN                      & 53.71          & 76.12          & \textbf{78.87}         & \textbf{93.16} \\ \midrule
  RANSAC(0.1)          & 57.18          & 79.16          & 74.12          & 91.37 \\
  RANSAC(0.2)          & 57.36          & 79.20          & 74.40          & 91.45 \\
  RANSAC(0.3)          & 57.27          & 79.20          & 74.13          & 91.35 \\
  RANSAC(0.4)          & 57.00          & 79.13          & 73.55          & 91.14 \\
  W-RANSAC(0.1) & 57.27          & 79.20          & 74.29          & 91.45 \\
  W-RANSAC(0.2) & 57.42          & \textbf{79.26} & 74.53 & 91.56 \\
  W-RANSAC(0.3) & 57.38          & 79.24          & 74.36          & 91.46 \\ \midrule
  pruned(0.75)         & \textbf{57.43} & 79.19          & 74.48          & 91.45 \\
  most confident    &   57.11	& 79.22	& 74.00	& 91.46
    \\ \bottomrule
  \end{tabular}\footnotesize
  RANSAC uses unit weights, while W-RANSAC is weighted by quaternion norm.
  PoseCNN and the best performing averaging methods are shown for comparison.
  Numbers in parentheses describe the clustering threshold in radians. Source:~\citet{capellen2020convposecnn}.
  \end{threeparttable}
  \end{table}

  \begin{table}[H]
    \centering
    \begin{threeparttable}
    \caption{Results for ConvPoseCNN Shape}
    \label{tab:convposecnnshaperesults}\setlength{\tabcolsep}{12pt}
    \begin{tabular}{lcccc}
    \toprule
            & \multicolumn{2}{c}{ 6D Pose }    & \multicolumn{2}{c}{Rotation only} \\
    \cmidrule(lr){2-3}\cmidrule(lr){4-5}
                          & AUC P          & AUC S          & AUC P           & AUC S    \\ \midrule
    PoseCNN  & 53.71 & 76.12     & \textbf{78.87} & \textbf{93.16}      \\ \midrule
    average & 54.27   & 78.94  & 70.02 &	90.91
    
              \\
    norm weighted    & \textbf{55.54} & 79.27  &  72.15 &	91.55  \\
    pruned(0.5)                & 55.33          & \textbf{79.29}  & 71.82 &	91.45 \\
    pruned(0.75)     & 54.62          & 79.09 & 70.56	 & 91.00  \\
    pruned(0.85)   & 53.86          & 78.85 & 69.34	& 90.57
     \\
    pruned(0.9)      & 53.23          & 78.66   & 68.37 & 90.25
       \\ \midrule
    RANSAC(0.2)                  & 49.44          & 77.65   & 63.09 & 	88.73   \\
    RANSAC(0.3)                  & 50.47          & 77.92       & 64.53	& 89.18      \\
    RANSAC(0.4)                  & 51.19          & 78.09      & 65.61	& 89.50 \\
    W-RANSAC(0.2) & 49.56  & 77.73  & 63.33 &	88.85 \\
    W-RANSAC(0.3)  & 50.54   & 77.91 & 64.78 & 	89.21 \\
    W-RANSAC(0.4) & 51.33  & 78.13  & 65.94	& 89.56 \\
    \bottomrule
    \end{tabular}
    \label{tab:resultsshapeconvposecnn}
    Table from \citet{capellen2020convposecnn}.
    \end{threeparttable}
    \end{table}

\subsubsection{Loss Variants}
The choice of aggregation method did not have a big impact on the models trained with $\text{QLoss}$ and thus we show only the results for $\text{Shape}$ variant in \cref{tab:convposecnnshaperesults}.
Among the averaging methods, norm weight improves the result, whereas pruning does not. This suggests that there are less-confident but important predictions with higher distance from the mean and removing them significantly affects the average. This could be an effect of training with the average quaternion, where such behavior is not discouraged. Both RANSAC variants---with and without weighting---resulted in comparatively worse results.
We conclude that the pixel-wise losses obtain superior performance, and average-before-loss scheme is not advantageous. Also, a fast dense version of SMLoss would need to be found in order to apply it in our architecture.

\subsubsection{ConvPoseCNN Final Results}

\begin{figure*}
\begin{tikzpicture}[font=\footnotesize]
\node[inner sep=0] (img) {\includegraphics[width=.97\textwidth]{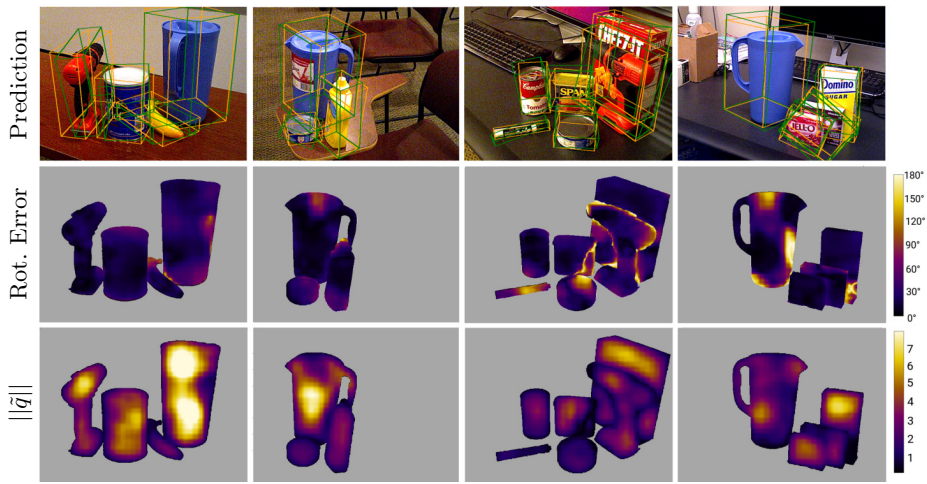}};

\node[anchor=south, rotate=90, text depth=0.25ex] at ($(img.north west)!0.16666!(img.south west)$) {Prediction};
\node[anchor=south, rotate=90, text depth=0.25ex] at ($(img.north west)!0.5!(img.south west)$) {Rot. Error};
\node[anchor=south, rotate=90, text depth=0.25ex] at ($(img.north west)!0.83333!(img.south west)$) {$||\tilde{q}||$};
\end{tikzpicture}
\caption{Qualitative results from ConvPoseCNN L2 on the YCB-Video test set.
Top: The orange boxes show the ground truth bounding boxes, the green boxes the 6D pose prediction.
Middle: Angular error of the dense quaternion prediction $\tilde{q}$ w.r.t. ground truth, masked by ground truth segmentation.
Bottom: Quaternion prediction norm $||\tilde{q}||$ before normalization. This measure is used for weighted aggregation.
Note that the prediction norm is low in high-error regions and high in
regions that are far from occlusions and feature-rich. Source:~\citet{capellen2020convposecnn}.}
\label{fig:qualitative_results}
\end{figure*}

  We start the discussion about ConvPoseCNN with the qualitative result shown in \cref{fig:qualitative_results}. We visualize the 3D ground truth and predictions for all the objects in the input scene as well as orientation error and predicted orientation norm per pixel. Dense pixel-wise orientation prediction makes it easier to visualize error at each pixel and to analyze them closely. A major observation from the visualizations is that the pixels in the feature-rich regions---close to object boundaries or distinctive textures---have lower orientation error while the pixels in the feature-poor regions exhibit higher angular error. A similar phenomenon is also observed in the prediction norm visualization. The pixels in feature-rich regions have higher norm orientation predictions while the pixels in feature-poor regions have lower norm orientation predictions.
  We hypothesize that in feature-rich regions, the network is confident of the predictions and thus the predictions are encouraged on one specific direction, whereas in the feature-poor regions, the predictions are pulled towards various possible directions resulting in predictions with a smaller norm.

\begin{table*}
\centering
\begin{threeparttable}
\caption{6D pose, translation, rotation, and segmentation results}
\label{tab:transresults}\footnotesize\setlength{\tabcolsep}{2pt}
\begin{tabular}{l@{\hspace{.2cm}}lcccccccc}
\toprule
                & & \multicolumn{2}{c}{ 6D pose }    & \multicolumn{2}{c}{Rotation only} &   NonSymC & SymC & Transl. & Segm. \\
\cmidrule(lr){3-4}\cmidrule(lr){5-6}\cmidrule(lr){7-7}\cmidrule(lr){8-8}\cmidrule(lr){9-9}\cmidrule(lr){10-10}
                &      & AUC P          & AUC S          & AUC P           & AUC S & AUC P           & AUC S  & Error [m] & IoU \\ \midrule

\parbox[t]{2mm}{\multirow{5}{*}{\rotatebox[origin=c]{90}{full network}}}

& PoseCNN                   & 53.71     & 76.12     & 78.87 & 93.16 & 60.49 &	63.28 & 0.0520  & 0.8369 \\
& PoseCNN\tnote{1} & 53.29 &	78.31
   & 69.00 &	90.49
 & 60.91 &	57.91
 & 0.0465 & 0.8071 \\
& ours, QLoss         & 57.16     & 77.08     & 80.51 & 93.35 & 64.75 & 53.95
& 0.0565 & 0.7725 \\
& ours, Shape         & 55.54     & 79.27     & 72.15 & 91.55 & 62.77 & 56.42
 & 0.0455 & 0.8038 \\
& ours, L2            & 57.42     & 79.26     & 74.53 & 91.56 & 63.48 & 58.85
 & 0.0411 & 0.8044 \\
\midrule
\parbox[t]{2mm}{\multirow{4}{*}{\rotatebox[origin=c]{90}{GT segm.}}}
& PoseCNN\tnote{1} & 52.90     & 80.11     & 69.60 & 91.63 & 76.63 & 84.15 & 0.0345 & 1      \\
& ours, QLoss         & 57.73     & 79.04     & 81.20 & 94.52 & 88.27 & 90.14 & 0.0386 & 1      \\
& ours, Shape         & 56.27     & 81.27     & 72.53 & 92.27 &  77.32  & 89.06 & 0.0316 & 1      \\
& ours, L2            & 59.50     & 81.54     & 76.37 & 92.32 & 80.67 & 85.52 & 0.0314 & 1      \\ \bottomrule
\end{tabular}\footnotesize
The average translation error, the segmentation IoU and the AUC metrics for different models. The AUC results were achieved using weighted RANSAC(0.1) for ConvPoseCNN QLoss, Markley's norm weighted average for ConvPoseCNN Shape and weighted RANSAC(0.2) for ConvPoseCNN L2.
\textit{GT segm.} refers to ground truth segmentation (i.e. only pose estimation).
Source:~\citet{capellen2020convposecnn}.
\begin{tablenotes}
 \item[1] Our own reimplementation.
\end{tablenotes}

\end{threeparttable}
\end{table*}

In \cref{tab:transresults}, we report the quantitative results of ConvPoseCNN models trained with three different loss functions---$\text{L2}$ and $\text{QLoss}$, and $\text{Shape}$--- and compare it with the PoseCNN baseline model provided in the YCB-Video Toolbox \footnote{\label{note:ycb_toolbox}\url{https://github.com/yuxng/YCB\_Video\_toolbox}}.

We provide AUC P and AUC S metric for all models including results from our own implementation of PoseCNN model in \cref{tab:transresults}. All the three variants of ConvPoseCNN perform slightly better than PoseCNN on both AUC P and AUC S metrics. Moreover, the ConvPoseCNN variant trained with $\text{L2}$ yields the best results among the ConvPoseCNN variants. $\text{QLoss}$ variant performed comparative to $\text{L2}$ variant on AUC P metric, whereas $\text{Shape}$ variant performed comparative to $\text{L2}$ variant on AUC S loss.

Moreover, to understand the influence of translation and orientation components on the overall AUC P and AUC S metric we report AUC P and AUC S metric computed for rotation only and translation error (computed in Meters) separately.
Although all the ConvPoseCNN variants perform slightly better than PoseCNN on the AUC metrics, only $\text{QLoss}$ variant performs better than PoseCNN on the rotation only AUC metrics. Analyzing the translation error suggests that the translation estimate influences the AUC losses more than the orientation estimate. However, the models that achieve better translation estimation, performs worse with the orientation estimate.

Furthermore, to analyze the performance of the models on objects with and without symmetry we report the average per-class AUC P metric for objects without symmetry and average per-class AUC S for objects with symmetry.
ConvPoseCNN performed a bit better than PoseCNN for the objects without symmetry but worse for the ones with symmetry. This can be explained by the use loss functions---QLoss and L2 loss---that are not designed to handle symmetry.
But, surprisingly, the model trained with SMLoss also performs worse for the symmetric objects compared to PoseCNN.

This might be due to different reasons: First, we utilize an average
before calculating the loss; therefore during training
the average might penalize predicting different shape-equivalent quaternions, in case their average is not
shape-equivalent. Secondly, there are only five symmetric objects in the dataset and we noticed that two
of those, the two clamp objects, are very similar and
thus challenging, not only for the orientation but as
well for the segmentation and vertex prediction. This
is further complicated by a difference in object coordinate systems for these two objects.

 While aggregating the dense pixel-wise orientation predictions to a single orientation prediction per-class, we use segmentation results. Thus, the segmentation results also influence the final metrics. To quantify the influence of segmentation results we report metrics for the all the five models---three ConvPoseCNN, and two PoseCNN variants---using the ground truth segmentation as well. Using ground truth segmentation improves translation and orientation for all the models. \citet{hu2019segmentation} also report a similar observation.

\begin{table}[H]
  \centering
  \begin{threeparttable}
  \caption{Comparison to Related Work}
  \label{tab:rwcomparison}\setlength{\tabcolsep}{12pt}
  \begin{tabular}{lccc}
  \toprule
                                       & \multicolumn{2}{c}{Total} &  Average \\
  \cmidrule(lr){2-3}\cmidrule(lr){4-4}
                                       & AUC P       & AUC S     & AUC \tnote{1}    \\ \midrule
  PoseCNN                      & 53.7        & 75.9        & 61.30  \\
  ConvPoseCNN L2 & 57.4        & 79.2         & 62.40  \\
  HeatMaps without FM      &            &                    & 61.41             \\ \midrule
  ConvPoseCNN+FM  & 58.22 & 79.55 & 61.59 \\
  HeatMaps with FM       &            &                    & 72.79     \\ \bottomrule
  \end{tabular}\footnotesize
  Comparison between PoseCNN (as reported by \citet{xiang2017posecnn}), ConvPoseCNN L2 with pruned(0.75), and HeatMaps \citep{heatmaps} without and with Feature Mapping (FM). Source:~\citet{capellen2020convposecnn}.
  \begin{tablenotes}
   \item[1] As defined by \citet{heatmaps}.
  \end{tablenotes}
  \end{threeparttable}
  \end{table}

  \begin{table}
    \begin{threeparttable}
    \vspace{.9em} %
    \caption{Detailed Class-wise Results}\label{tab:bestconvposecnn}\setlength{\tabcolsep}{12pt}
    \begin{tabular}{lrrrrr} \toprule
    Class                                                  & \multicolumn{2}{c}{Ours} & \multicolumn{2}{c}{PoseCNN} \\
    \cmidrule(lr){2-3}\cmidrule(lr){4-5}
                                                           & AUC P        & AUC S  &  AUC P       & AUC S \\
    \midrule
    master$\_$chef$\_$can                           & 62.32        & 89.55  &  50.08       & 83.72 \\
    cracker$\_$box                                  & 66.69        & 83.78  &  52.94       & 76.56 \\
    sugar$\_$box                                    & 67.19        & 82.51  &  68.33       & 83.95 \\
    tomato$\_$soup$\_$can                           & 75.52        & 88.05  &  66.11       & 80.90 \\
    mustard$\_$bottle                               & 83.79        & 92.59  &  80.84       & 90.64 \\
    tuna$\_$fish$\_$can                             & 60.98        & 83.67  &  70.56       & 88.05 \\
    pudding$\_$box                                  & 62.17        & 76.31  &  62.22       & 78.72 \\
    gelatin$\_$box                                  & 83.84        & 92.92  &  74.86       & 85.73 \\
    potted$\_$meat$\_$can                           & 65.86        & 85.92  &  59.40       & 79.51 \\
    banana                                          & 37.74        & 76.30  &  72.16       & 86.24 \\
    pitcher$\_$base                                 & 62.19        & 84.63  &  53.11       & 78.08 \\
    bleach$\_$cleanser                              & 55.14        & 76.92  &  50.22       & 72.81 \\
    bowl                    & 3.55         & 66.41  &  3.09        & 70.31 \\
    mug                                             & 45.83        & 72.05  &  58.39       & 78.22 \\
    power$\_$drill                                  & 76.47        & 88.26  &  55.21       & 72.91 \\
    wood$\_$block           & 0.12         & 25.90  &  26.19       & 62.43 \\
    scissors                                        & 56.42        & 79.01  &  35.27       & 57.48 \\
    large$\_$marker                                 & 55.26        & 70.19  &  58.11       & 70.98 \\
    large$\_$clamp          & 29.73        & 58.21  &  24.47       & 51.05 \\
    extra$\_$large$\_$clamp & 21.99        & 54.43  &  15.97       & 46.15 \\
    foam$\_$brick           & 51.80        & 88.02  &  39.90       & 86.46 \\
    \bottomrule
    \end{tabular}
    Source:~\citet{capellen2020convposecnn}.
    \end{threeparttable}
    \end{table}

\subsubsection{Comparison to Related Work}

In Table \ref{tab:rwcomparison}, we show the comparisons between ConvPoseCNN, PoseCNN, and HeatMaps \citep{heatmaps} approaches. \citet{heatmaps} report class-wise area under the accuracy curve metric (AUC) instead of AUC P and AUC S metrics. To make the methods comparable, we provide AUC for both ConvPoseCNN and PoseCNN.
\citep{heatmaps} proposed Feature Mapping (FM) technique that significantly improves their results. Without feature mapping, we perform slightly better than both PoseCNN and HeatMaps. However, the difference is negligible considering the variations due to the choice of hyperparameters and minor implementations details.
Detailed class-wise AUC metrics for both the best performing ConvPoseCNN and PoseCNN models are shown in \ref{tab:bestconvposecnn}.

We also investigated applying the Feature Mapping technique \citep{heatmaps} to our model.
Following the process, we render synthetic images with poses corresponding to the real training data. We selected the features from backbone VGG-16 for the mapping process and thus have to transfer two feature
maps with 512 features each. We replaced the fullyconnected network architecture for feature mapping as done by \citep{heatmaps}, with a convolutional set-up and mapped the feature from the different stages to each other with residual blocks based on (1×1) convolutions. The results are presented in \ref{tab:rwcomparison}. 
However, we did not observe the large gains reported by \citep{heatmaps} for our architecture. We hypothesize that the feature mapping technique is highly dependent on the quality and distribution of the rendered synthetic images, which are maybe not of sufficient quality in our case.

\subsubsection{Time Comparisons}

\begin{table}[H]
\centering
\begin{threeparttable}
\caption{Training performance \& model sizes}
\label{tab:timecomparisontrain}\setlength{\tabcolsep}{12pt}
\begin{tabular}{lcl} \toprule
                       & Iterations/s\tnote{1} & Model size \\ \midrule
PoseCNN                & 1.18          & 1.1 GiB     \\
ConvPoseCNN L2         & 2.09          & 308.9 MiB   \\
ConvPoseCNN QLoss      & 2.09          & 308.9 MiB   \\
ConvPoseCNN SMLoss & 1.99          & 308.9 MiB   \\ \bottomrule
\end{tabular}
\footnotesize
\begin{tablenotes}
 \item [1] Using a batch size of 2. Averaged over 400 iterations.
\end{tablenotes}
Source:~\citet{capellen2020convposecnn}.
\end{threeparttable}
\end{table}

We used NVIDIA GTX 1080 Ti GPU with 11\,GB of memory to benchmark the training and inference time for ConvPoseCNN and PoseCNN models. In table \ref{tab:timecomparisontrain} we report number of iterations per second. All the variants of ConvPoseCNN are significantly faster. Additionally, size of the saved ConvPoseCNN models are significantly smaller compared to the PoseCNN models.

Unfortunately, this advantage in speed during the training process is not observed during the inference as shown in \ref{tab:testtime}. Averaging methods, on average, consume time comparable to the PoseCNN. But the RANSAC based clustering methods more time consuming; the forward pass of ConvPoseCNN takes about 65.5\,ms, the Hough transform around 68.6\,ms. We attribute the comparable inference time consumption to the highly optimized ROI pooling layers in the modern deep learning frameworks. 

\begin{table}
\centering
\begin{threeparttable}
\caption{Inference timings}
\label{tab:testtime}\setlength{\tabcolsep}{12pt}
\begin{tabular}{lrr} \toprule
Method                    & Time [ms]\tnote{1}  & Aggregation [ms] \\ \midrule
PoseCNN\tnote{2} & 141.71 &                  \\ \midrule
ConvPoseCNN & & \\
- naive average         & 136.96 & 2.34             \\
- average               & 146.70 & 12.61            \\
- weighted average      & 146.92 & 13.00            \\
- pruned w. average & 148.61 & 14.64            \\
- RANSAC                & 158.66 & 24.97 \\
- w. RANSAC       & 563.16 & 65.82         \\ \bottomrule
\end{tabular}
\footnotesize
\begin{tablenotes}
 \item [1] Single frame, includes aggregation.
 \item [2] \citet{xiang2017posecnn}.
\end{tablenotes}
Source:~\citet{capellen2020convposecnn}.
\end{threeparttable}
\end{table}

\subsubsection{Iterative Refinement}

Post-prediction iterative refinement module is trained with segmentation, translation, and orientation estimates from ConvPoseCNN as well as VGG16 features as input.
At each iteration, the model refines segmentation, translation, and orientation estimates. VGG16 features provide contextual information about the input scene.
We experimented with varying number of refinement steps. Similar to ConvPoseCNN, we used same combined loss function as discussed in \cref{sub:training}.
But, we observed both training and validation loss plateauing very early on the training process and the resulting model also performed worse quantitatively compared to ConvPoseCNN on the test set.

This could be because the estimates are in a form that is not a suitable for a simple three layer network. Exploring complex architectures is not an option for us since we focus on keeping the overhead of iterative refinement minimal.

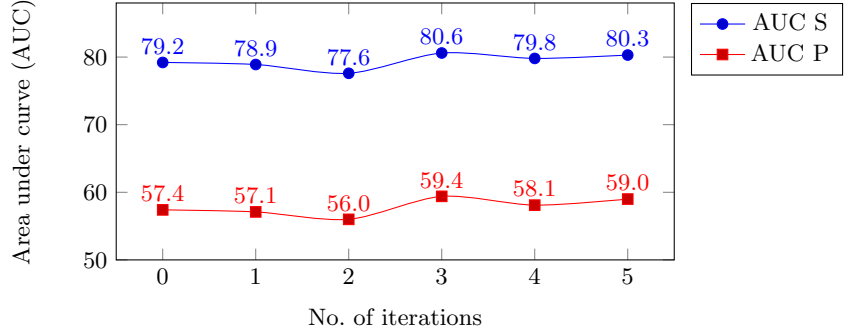
\begin{figure}
  \centering
﻿    \begin{tikzpicture}
\begin{axis}[
    xlabel=No. of  iterations,
    ylabel=Area under curve (AUC),
    ymin=50, ymax=88,
    xtick={0,  1, 2, 3, 4, 5},
    ytick={0,10,...,100},
    width=9cm,
    height=5cm,
    legend pos=outer north east,
    nodes near coords style={/pgf/number format/.cd,fixed zerofill,precision=1},
    nodes near coords,
            ]

\addplot+[smooth,mark=*,blue] plot coordinates {
 (0,     79.2)
 (1,     78.9)
 (2,    77.6)
 (3,    80.6)
 (4,    79.8)
 (5,   80.3)
};

\addplot+[smooth,color=red]
    plot coordinates {
 (0,    57.4)
 (1,     57.1)
 (2,    56.0)
 (3,    59.4)
 (4,    58.1)
 (5,   59.0)
    };
\addlegendentry{AUC S}
\addlegendentry{AUC P}
\end{axis}
 \end{tikzpicture}
   \caption{Results of pre-prediction feature refinement process for various number of iterations. The variant with zero iterations corresponds to ConvPoseCNN without any refinement (\cref{tab:rwcomparison}).}
  \label{fig:tikz:ir_results}
\end{figure}

\begin{table}
  \vspace{.9em} %
  \caption{Class-wise Results ConvPoseCNN without refinement, with three iterations of refinement and with five iterations of refinement.}\label{tab:ir_class}\setlength{\tabcolsep}{8pt}
  \begin{tabular}{lrrrrrrr} \toprule
  Class                                                  & \multicolumn{2}{c}{PoseCNN} & \multicolumn{2}{c}{IR 3} & \multicolumn{2}{c}{IR 5} \\
  \cmidrule(lr){2-3}\cmidrule(lr){4-5}\cmidrule(lr){6-7}
                                                         & AUC P        & AUC S  &  AUC P       & AUC S &  AUC P       & AUC S \\
  \midrule
  master$\_$chef$\_$can                           & 62.32        & 89.55  &  62.69       &  90.93      &     61.58  &     91.09        \\
  cracker$\_$box                                  & 66.69        & 83.78  &  51.64       &  79.02      &     62.48  &     82.53        \\
  sugar$\_$box                                    & 67.19        & 82.51  &  63.16       &  80.81      &     68.95  &     84.16        \\
  tomato$\_$soup$\_$can                           & 75.52        & 88.05  &  78.70       &  90.70      &     75.12  &     88.65        \\
  mustard$\_$bottle                               & 83.79        & 92.59  &  83.66       &  92.09      &     83.99  &     91.65       \\
  tuna$\_$fish$\_$can                             & 60.98        & 83.67  &  71.10       &  88.15      &     72.68  &     90.37        \\
  pudding$\_$box                                  & 62.17        & 76.31  &  67.72       &  84.73      &     66.11  &     83.25        \\
  gelatin$\_$box                                  & 83.84        & 92.92  &  83.38       &  91.45      &     86.98  &     93.18       \\
  potted$\_$meat$\_$can                           & 65.86        & 85.92  &  69.52       &  87.56      &     68.21  &     86.22        \\
  banana                                          & 37.74        & 76.30  &  42.96       &  70.24      &     42.75  &     70.34        \\
  pitcher$\_$base                                 & 62.19        & 84.63  &  68.31       &  86.79      &     66.51  &     86.55        \\
  bleach$\_$cleanser                              & 55.14        & 76.92  &  50.86       &  71.48      &     52.28  &     75.61        \\
  bowl                                            & 3.55         & 66.41  &  7.21        &  73.04      &     8.24   &     69.29       \\
  mug                                             & 45.83        & 72.05  &  58.31       &  81.68      &     62.11  &     82.64        \\
  power$\_$drill                                  & 76.47        & 88.26  &  73.12       &  86.57      &     71.60  &     85.68        \\
  wood$\_$block                                   & 0.12         & 25.90  &  0.785       &  27.70      &     1.07   &     31.86       \\
  scissors                                        & 56.42        & 79.01  &  62.41       &  80.96      &     51.22  &     75.56        \\
  large$\_$marker                                 & 55.26        & 70.19  &  64.16       &  76.35      &     60.15  &     71.96        \\
  large$\_$clamp                                  & 29.73        & 58.21  &  35.66       &  62.34      &     33.14  &     61.93        \\
  extra$\_$large$\_$clamp                         & 21.99        & 54.43  &  23.16       &  55.74      &     23.91  &     55.91        \\
  foam$\_$brick                                   & 51.80        & 88.02  &  51.31       &  88.62      &     47.69  &     87.08        \\
  \bottomrule
  \end{tabular}
  \end{table}

In contrast to the post-prediction refinement, pre-prediction refinement not only performed well during but also improved the AUC metrics on the test set. This suggests that in the case of ConvPoseCNN, refining the features at an early stage helps the network in learning representations better suitable for pose estimation.
We trained the refinement module with a various number of iterations and in \cref{fig:tikz:ir_results}, we present the AUC metrics
achieved by various number of refinement iterations. Overall, the iterative refinement improves the prediction and different number of iterations results in slightly different AUC metrics.
Interestingly, the performance peaks at three iterations. If there are any gains with
more iterations, they are not significant. We attribute this fact to the small
depth of our refinement network which limits the operations it can perform.
In \cref{tab:ir_class} we compare the class-wise AUC metrics for ConvPoseCNN (without refinement), three and five iterations of refinement. For most objects, the AUC metrices are improved but for some objects, there is a drop in accuracy. The maximum gain of 12.48 AUC-P and 9.63 is observed for the \textit{mug} object while a severe drop of 15.05 AUC-P and 5.54 is observed for \textit{cracker\_box} and \textit{bleach\_cleanser} respectively---both relatively big objects, where information needs to be communicated and fused over larger regions.

\section{Conclusion}
We presented ConvPoseCNN, a fully convolutional architecture for object pose estimation and demonstrated that, similar to translation estimation, direct regression of the orientation estimation can be done in a dense pixel-wise manner. This helps in not only simplifying neural networks architectures for $\text{6D}$ object pose estimation but also reducing the size of the models and faster training.
To further the performance of fully convolutional models for pose estimation, scalable dense pixel-wise loss function needs to be explored. 
As a next step, we plan to evaluate ConvPoseCNN on highly cluttered scenes where we expect the dense predictions to be especially beneficial, since disambiguation of close objects should be more direct than with RoI-based architectures.

Moreover, we demonstrated that the pose predictions can be refined even with a small network to boost the performance, provided the refinement is done at the right level of abstraction.
In case of ConvPoseCNN, refining intermediate representations yielded better performance than post-prediction refinement. Thus, network architecture designs that imbue refinement modules should be favoured for object pose estimation. In the future, we plan to combine iterative refinement with other state-of-the-art architectures and further investigate the design of refinement modules. The key challenge is to balance the need for a larger powerful module capable of iterative refinement with keeping the processing time and memory overhead introduced by the refinement module low.

\bibliography{references}
\end{document}